\documentclass[11pt]{article}

\usepackage[final]{acl}

\usepackage{latexsym}
\usepackage{booktabs}
\usepackage{amsmath}
\usepackage[T1]{fontenc}
\usepackage{fontspec}

\newfontfamily\myanmarfont[
  Path=fonts/noto/,
  Extension=.ttf,
  UprightFont=NotoSansMyanmar-Regular,
  Scale=1.1
]{Noto Sans Myanmar}

\newfontfamily\arabicfont[
  Path=fonts/noto/,
  Extension=.ttf,
  UprightFont=NotoNaskhArabic-Regular,
  Script=Arabic,
  Scale=1.1
]{Noto Naskh Arabic}

\newcommand{\textmyanmar}[1]{{\myanmarfont #1}}
\newcommand{\textarabic}[1]{{\arabicfont #1}}

\usepackage{lipsum}
\usepackage{tipa}
\let\origtextipa\textipa
\renewcommand{\textipa}[1]{{\rmfamily\origtextipa{#1}}}
\usepackage{array}

\usepackage{microtype}
\usepackage{hyperref}

\usepackage{inconsolata}

\usepackage{graphicx}

\title{Where Are We At with Automatic Speech
Recognition for the Bambara Language?}

\author{
\textbf{Seydou Diallo}$^{1,4,5}$,
\textbf{Yacouba Diarra}$^{1,2}$,
\textbf{Mamadou K. KEITA}$^{1,3}$,\\
\textbf{Panga Azazia Kamaté}$^{1,2}$,
\textbf{Adam Bouno Kampo}$^{1}$,
\textbf{Aboubacar Ouattara}$^{4}$\\
\\
$^{1}$MALIBA-AI \quad
$^{2}$RobotsMali AI4D Lab \quad
$^{3}$Rochester Institute of Technology \quad
$^{4}$DJELIA\\
$^{5}$Dakar American University of Science and Technology
}

\begin{document}
\maketitle
\begin{abstract}
This paper introduces the first standardized benchmark for evaluating Automatic Speech Recognition (ASR) in the Bambara language, utilizing one hour of professionally recorded Malian constitutional text. Designed as a controlled reference set under near-optimal acoustic and linguistic conditions, the benchmark was used to evaluate 37 models, ranging from Bambara-trained systems to large-scale commercial models. Our findings reveal that current ASR performance remains significantly below deployment standards in a narrow formal domain; the top-performing system in terms of Word Error Rate (WER) achieved 46.76\% and the best Character Error Rate (CER) of 13.00\% was set by another model, while several prominent multilingual models exceeded 100\% WER. These results suggest that multilingual pre-training and model scaling alone are insufficient for underrepresented languages. Furthermore, because this dataset represents a best-case scenario of the most simplified and formal form of spoken Bambara, these figures are yet to be tested against practical, real-world settings. We provide the benchmark and an accompanying public leaderboard to facilitate transparent evaluation and future research in Bambara speech technology.
\end{abstract}

\section{Introduction}
\label{sec:intro}

\begin{figure*}[ht]
    \centering
    \includegraphics[width=\linewidth]{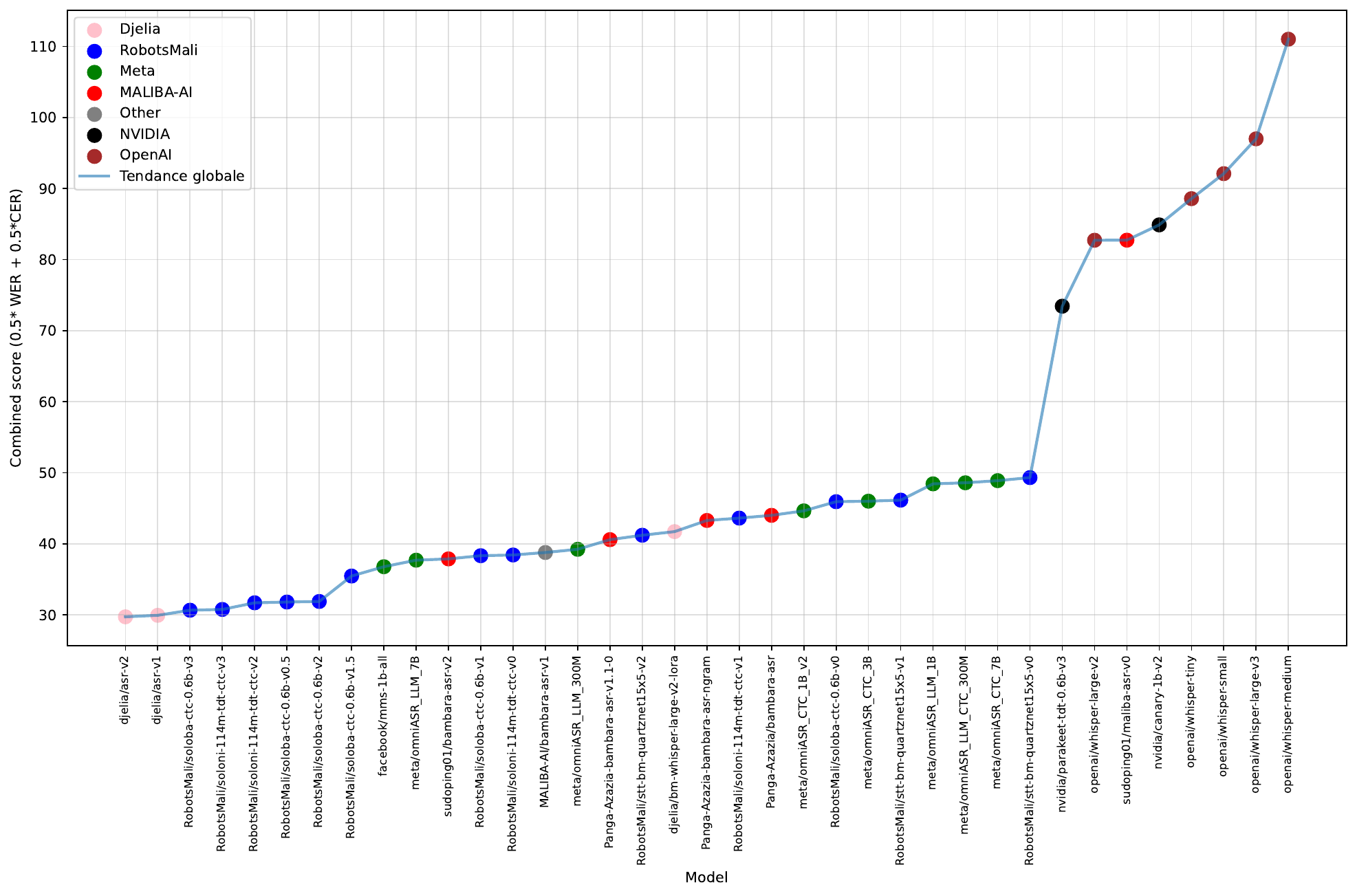}
    \caption{Models combined performance on Bambara Benchmark. Lower is better.}
    \label{fig:data-efficiency}
\end{figure*}

Automatic Speech Recognition (ASR) for Bambara has seen growing interest in the past three years. Since the 2022 release of Jeli-ASR \cite{Diarra2022Griots}, the first open ASR dataset for the language, numerous models and datasets have emerged from both research labs and community initiatives. However, this rapid growth raises concerns about quality and usability, concerns that cannot be addressed without standardized evaluation.

Quality, when it comes to low resource African languages, is the object of strong debates among the African NLP community due to the variety of dialects, writing systems, and standards \cite{hussen2025statelargelanguagemodels}, but also the complexity of the contact phenomenon between African languages and western languages, namely code switching.

As the Word Error Rate (WER) is only relevant when we have already defined and assessed the quality of the evaluation set, Whatever quality means for one, some researchers recommend defaulting to human evaluation by native speakers (\citealp{lau2025dataqualityissuesmultilingual}; \citealp{tall_2025_17672774}). However, this process is time consuming and expensive, furthermore edit distance metrics like WER or Character Error Rate (CER) remain insightful on a curated and standardized benchmark.

However, no such benchmark existed for evaluating Bambara ASR models, most openly released models\footnote{\href{https://hf.co/facebook}{hf.co/facebook}; \href{https://github.com/facebookresearch/omnilingual-asr}{facebookresearch/omnilingual-asr}; \href{https://hf.co/asr-africa}{hf.co/asr-africa}; \href{https://hf.co/MALIBA-AI}{hf.co/MALIBA-AI}; \href{https://hf.co/RobotsMali}{hf.co/RobotsMali}; \href{https://hf.co/djelia}{hf.co/djelia}} report values for WER and CER on internal test sets. To address this issue and offer a \textit{reference test set}, we publish the first Bambara ASR benchmark and leaderboard backed with experts validated transcriptions.

As more data collection initiatives for African languages emerge, often with strict rules to capture simplified language and context, such as no slang, no code-switching, no background noise etc, we have designed this first benchmark to represent an equally "pure" version of the Bambara language. Relatively poor evaluation results of models trained on more modern and accessible Bambara (see section \ref{sec:eval}) raise questions about the representativeness and usability of simplified language for real-world applications where natural data often include noise, informal terms, and code-switching. Therefore, we anticipate that this benchmark will be among the most difficult test sets for current Bambara ASR systems, covering a specialized and highly formal domain, and we argue for its interpretation as a reference test set for \textit{pure Bambara}.

\section{Characteristics of the Benchmark}
\label{sec:dataset}
This first version of the evaluation set consists of a 1 hour recording of a professionally translated version of the Malian constitution, translated and recorded by the Direction Nationale de l'Education Non Formelle et des Langues Nationales (DNENF-LN)\footnote{DNENF-LN is the government founded organization in charge of literacy training and official documents translation in all the 13 national languages of Mali: https://dnenfln.ml/} under studio conditions, featuring one unique adult male voice.

With the premier legal text of Mali as topic, the dataset features a highly formal and diverse vocabulary that unpacks many aspects of the organization of Malian society, laws, institutions, rights and responsibilities, all written in the Bambara latin script using standard orthography and \textbf{without code switching}. The dataset also has an important representation of numbers, as the constitution contains 191 articles as of July 2023, 160 of which are clearly spelled out in the recording, specifically in ordinal forms. 

We ran manual segmentation and audio-text alignment using the Audacity software \cite{Audacity}. Then we performed a final quality assurance step wherein the aligned utterances were reviewed to correct divergences resulting from the corpus' read speech (READ) nature, specifically addressing instances where the speaker paraphrased or interpreted the text rather than providing a literal recital. This process resulted in 500 variable-length audio utterances ranging from \emph{600 ms} to \emph{46 seconds}, with a mean duration of \emph{7.57 seconds}. With this variability the benchmark aims to test models' capabilities on both short and long form transcription.

We calculated Signal-to-noise Ratio (SNR) as an estimate of the acoustic purity of the benchmark (a higher value is best). We used the same Voice-activity-detection based implementation and classification thresholds as \citeauthor{diarra2025dealinghardfactslowresource} but we calculated SNR on the segmented utterances, because the original recording features transition music and longer silences that would hinder the accuracy of the estimation as any non-speech segment is considered for estimating the Noise Power in this implementation (\citealp{diarra2025dealinghardfactslowresource}; \citealp{vad_snr}). Note that we still kept 8 of these silent/music segments in the final benchmark to test the robustness of the models, especially the tendency to "hallucinate" tokens as the silence becomes lengthy when there is in fact no speech. Table \ref{tab:snr} shows the SNR distribution the 492 remaining speech segments.

\begin{table}[!ht]
    \centering
    \begin{tabular}{lcc}
    \toprule
    \textbf{SNR Category} & \textbf{Threshold (dB)} & \textbf{Recordings} \\
    \midrule
    Medium SNR & $[5, 15)$ & 5 \\
    High SNR & $[15, 25)$ & 109 \\
    Very high SNR & $\ge 25$ & 378 \\
    \midrule
    \multicolumn{2}{r}{\textbf{Total Audios}} & \textbf{492} \\
    \bottomrule
    \end{tabular}
    \caption{Distribution of Audio utterances by Signal-to-noise Ratio Category.}
    \label{tab:snr}
\end{table}

We note that 99\% of the utterances are classified as relatively noise-free. This is an important point for interpreting our results: \textbf{this benchmark represents near-optimal acoustic conditions}\footnote{In future versions, we will collect data in various domains under different recording conditions, trying to maximize diversity and real world representativeness instead of purity}. Any production deployment would face significantly more challenging audio quality, so these results should be interpreted with caution given the specialized and nature of the reference test set and its acoustic purity.


\section{Leaderboard and Results of Open Bambara ASR Models}
\label{sec:eval}
We evaluated 37 publicly available ASR models on our benchmark, including monolingual ASR models, multilingual models with Bambara support, and large-scale commercial ASR systems. Table \ref{tab:results} presents the complete leaderboard ranked by a weighted average score of WER and CER (50\% WER + 50\% CER). This equal weighting reflects a neutral stance that does not privilege either word-level or character-level accuracy, treating both as equally informative for assessing transcription quality. We acknowledge that optimal weighting may depend on downstream application requirements for instance, applications sensitive to semantic accuracy may prioritize WER, while those tolerant of word boundary errors may favor CER. To address this, our public leaderboard allows users to adjust these weights according to their specific needs, and we report sensitivity analysis under alternative weightings in Table \ref{tab:sensitivity}. All evaluations were conducted using normalized text (lowercase, no punctuation \& consecutive whitespace) to ensure fair comparison between models.

\begin{table*}[!ht]
    \centering
    \small
    \begin{tabular}{llcccl}
    \toprule
    \textbf{Rank} & \textbf{Model} & \textbf{WER (\%)}$\downarrow$ & \textbf{CER (\%)}$\downarrow$ & \textbf{Combined (\%)}$\downarrow$ & \textbf{License} \\
    \midrule
    1 & djelia/asr-v2 & 47.50 & 13.56 & 29.73 & Proprietary \\
    2 & djelia/asr-v1 & 48.56 & 13.00 & 29.94 & Proprietary \\
    3 & RobotsMali/soloba-ctc-0.6b-v3 & 46.76 & 16.02 & 30.66 & Open Source \\
    4 & RobotsMali/soloni-114m-tdt-ctc-v3 & 48.32 & 14.81 & 30.77 & Open Source \\
    5 & RobotsMali/soloni-114m-tdt-ctc-v2 & 49.42 & 15.58 & 31.70 & Open Source \\
    6 & RobotsMali/soloba-ctc-0.6b-v0.5 & 49.93 & 15.33 & 31.81 & Open Source \\
    7 & RobotsMali/soloba-ctc-0.6b-v2 & 48.06 & 17.19 & 31.89 & Open Source \\
    8 & RobotsMali/soloba-ctc-0.6b-v1.5 & 52.56 & 19.93 & 35.47 & Open Source \\
    9 & facebook/mms-1b-all & 61.06 & 14.71 & 36.78 & Open Source \\
    10 & meta/omniASR\_LLM\_7B & 62.57 & 15.08 & 37.70 & Open Source \\
    11 & sudoping01/bambara-asr-v2 & 60.33 & 17.46 & 37.88 & Open Source \\
    12 & RobotsMali/soloba-ctc-0.6b-v1 & 57.59 & 20.81 & 38.33 & Open Source \\
    13 & RobotsMali/soloni-114m-tdt-ctc-v0 & 55.79 & 22.65 & 38.43 & Open Source \\
    14 & MALIBA-AI/bambara-asr-v1 & 61.74 & 17.90 & 38.78 & Open Source \\
    15 & meta/omniASR\_LLM\_300M & 63.32 & 17.32 & 39.23 & Open Source \\
    16 & Panga-Azazia/bambara-asr-v1.1-0 & 60.39 & 22.60 & 40.59 & Proprietary \\
    17 & RobotsMali/stt-bm-quartznet15x5-v2 & 65.66 & 18.98 & 41.21 & Open Source \\
    18 & djelia/bm-whisper-large-v2-lora & 59.17 & 25.85 & 41.72 & Proprietary \\
    19 & Panga-Azazia/bambara-asr-ngram & 69.13 & 19.80 & 43.29 & Open Source \\
    20 & RobotsMali/soloni-114m-tdt-ctc-v1 & 61.14 & 27.69 & 43.62 & Open Source \\
    21 & Panga-Azazia/bambara-asr & 70.00 & 20.39 & 44.01 & Open Source \\
    22 & meta/omniASR\_CTC\_1B\_v2 & 69.62 & 21.93 & 44.64 & Open Source \\
    23 & RobotsMali/soloba-ctc-0.6b-v0 & 62.93 & 30.48 & 45.93 & Open Source \\
    24 & meta/omniASR\_CTC\_3B & 72.62 & 21.80 & 46.00 & Open Source \\
    25 & RobotsMali/stt-bm-quartznet15x5-v1 & 72.98 & 21.75 & 46.15 & Open Source \\
    26 & meta/omniASR\_LLM\_1B & 78.31 & 21.29 & 48.44 & Open Source \\
    27 & meta/omniASR\_LLM\_CTC\_300M & 76.87 & 22.87 & 48.59 & Open Source \\
    28 & meta/omniASR\_CTC\_7B & 74.65 & 25.47 & 48.89 & Open Source \\
    29 & RobotsMali/stt-bm-quartznet15x5-v0 & 75.82 & 25.23 & 49.32 & Open Source \\
    30 & nvidia/parakeet-tdt-0.6b-v3 & 100.06 & 49.24 & 73.44 & Open Source \\
    31 & openai/whisper-large-v2 & 106.84 & 60.80 & 82.72 & Open Source \\
    32 & sudoping01/maliba-asr-v0 & 94.86 & 71.72 & 82.73 & Open Source \\
    33 & nvidia/canary-1b-v2 & 111.64 & 60.55 & 84.88 & Open Source \\
    34 & openai/whisper-tiny & 112.72 & 66.61 & 88.57 & Open Source \\
    35 & openai/whisper-small & 109.97 & 75.84 & 92.09 & Open Source \\
    36 & openai/whisper-large-v3 & 121.06 & 75.10 & 96.99 & Open Source \\
    37 & openai/whisper-medium & 123.18 & 99.95 & 111.01 & Open Source \\
    \bottomrule
    \end{tabular}
    \caption{Bambara ASR Benchmark Leaderboard. Combined Score = 0.5 $\times$ WER + 0.5 $\times$ CER. Lower scores indicate better performance.}
    \label{tab:results}
\end{table*}

\subsection{Assessment}

The main finding of this evaluation is that current Bambara ASR systems do not yet meet the commonly accepted production-readiness thresholds in the narrow domain represented in our test set. Under our combined evaluation metric, the highest-ranked model attains a Word Error Rate of 47.50\%, indicating that nearly half of all words are incorrectly transcribed.

For context, production-grade ASR systems for well-resourced languages typically achieve Word Error Rates in the 5--15\% range \cite{nahabwe2025benchmarking}. Current Bambara ASR performance therefore remains approximately 30--40 percentage points below these levels, suggesting a substantial gap that will require significant advances in data, modeling, and evaluation to close.

Real-world Bambara speech introduces additional challenges: phone-quality or ambient recordings, multiple speakers with varying accents and dialects, ubiquitous French code-switching, informal vocabulary, variable recording equipment, and background noise. Therefore, this benchmark gives little insight into the performance of these models with truly naturalistic speech.

\subsection{Model-Specific Findings}
\label{subsec:model-specific}
We find that specialized fine-tunes from Djelia and RobotsMali substantially outperform their base version (parakeet, whisper) and all the other models from large multilingual initiatives. 

\paragraph{Multilingual models exhibit high error rates.} All evaluated OpenAI Whisper variants exhibit WER exceeding 100\%, indicating that models generate more tokens than present in the reference audio, a hallucination phenomenon. This pattern is consistent across model sizes: whisper-tiny (112.72\%), whisper-small (109.97\%), whisper-medium (123.18\%), whisper-large-v2 (106.84\%), and whisper-large-v3 (121.06\%). NVIDIA's Parakeet-tdt-0.6b-v3 (100.06\% WER) and Canary-1b-v2 (111.64\% WER) show similar behavior.

These results are consistent with findings that off-the-shelf multilingual ASR models require language-specific adaptation to perform well in underrepresented languages \cite{nahabwe2025benchmarking}. It is important to note that, while multilingual, the base versions of Whisper and Canary, along with Nvidia's monolingual Parakeet models, included in this study, did not include Bambara in respective their training sets. However, evaluating them allowed us to rule out the hypothesis that massive multilingualism may translate to better performance on unseen, underrepresented African languages like Bambara through transfer learning. On the other end, remarkably better performance from Meta's Omnilingual ASR and MMS models shows that even a negligible amount of Bambara data in the training set can drastically change these figures. 

\paragraph{Model scale does not compensate for data scarcity.} Meta's omniASR family provides insight into scaling effects. The 7B parameter CTC model (74.65\% WER) performs worse than the 300M LLM variant (63.32\% WER), and both lag behind the 114M parameter monolingual \texttt{soloni} models (48.32\% WER).

\paragraph{Character-level accuracy exceeds word-level accuracy.} CER results are notably better than WER across all models, with the best achieving 13.00\% (djelia/asr-v1). This suggests that models capture phonetic patterns more successfully than word boundaries and vocabulary, a pattern consistent with the challenges of morphologically rich languages where compound words and agglutination are frequent.

\subsection{Qualitative Error Analysis}

To better illustrate model failure modes, we present representative examples from our evaluation.

\paragraph{Hallucination in multilingual models.}
Table~\ref{tab:hallucination} shows severe hallucination in Whisper models, where the output contains scripts entirely unrelated to Bambara.   

\begin{table}[ht]
\centering
\small
\begin{tabular}{p{2.5cm}p{4.5cm}}
\toprule
\textbf{Model} & \textbf{Output (Audio ID: 85)} \\
\midrule
Reference & Baaras\textipa{O}r\textipa{O} ni s\textipa{E}g\textipa{E}nnafi\textipa{\textltailn}\textipa{E}b\textipa{O}s\textipa{O}r\textipa{O}li hak\textipa{E}w lakod\textipa{O}nnen don wa b\textipa{E}\textipa{E} n'u ka kan. \\
\midrule
whisper-medium & \textmyanmar{ခေေေေေေေေေေေေေေေ}... (Myanmar script) \\
whisper-large-v2 & \textarabic{بارا سروني سيڭنافين بو سروني حڭاو}... (Arabic script) \\
whisper-small & \textarabic{بارا صراني سيقنا في نبوصر}... (Arabic script) \\
\bottomrule
\end{tabular}

\caption{Hallucination in multilingual models: generate tokens from a different, unrelated language.}
\label{tab:hallucination}
\end{table}

\paragraph{Word boundary errors and morphological complexity.}
Table~\ref{tab:word-boundary} illustrates cases where character-level accuracy is high but word-level accuracy is low. This pattern reflects challenges with Bambara's agglutinative morphology, where compound words are common.



\begin{table}[ht]
\centering
\footnotesize
\setlength{\tabcolsep}{4pt}
\begin{tabular}{@{}p{5.5cm}rr@{}}
\toprule
\textbf{Reference} & \textbf{WER} & \textbf{CER} \\
\midrule
Dakun filanan: jamana ka y\textipa{E}r\textipa{E}mah\textipa{O}r\textipa{O}nya. & 40.0\% & 2.8\% \\[3pt]
Jamana sago b'a ka nafolomaf\textipa{E}nw n'a ka dugukolo s\textipa{O}r\textipa{O}f\textipa{E}nw b\textipa{E}\textipa{E} la. & 36.4\% & 3.3\% \\[3pt]
Tilay\textipa{O}r\textipa{O} duurunan: kiiritig\textipa{E}fanga. Dakun filanan jamana kiiritig\textipa{E}bulonba. & 85.7\% & 5.9\% \\
\bottomrule
\end{tabular}
\caption{Word boundary errors: low CER but high WER due to compound word segmentation.}
\label{tab:word-boundary}
\end{table}

For example, in utterance 125, the reference contains the compound \textit{y\textipa{E}r\textipa{E}mah\textipa{O}r\textipa{O}ya} (``Sovereignty/Independence''), which the MMS model segments as \textit{y\textipa{E}r\textipa{E}ma h\textipa{O}r\textipa{O}ya} nearly identical at the character level but counted as two word errors. Similarly, utterance 450 contains the compound: \textit{kiritig\textipa{E}fanga} (``judicial power'') becomes \textit{kiritig\textipa{E} fanga}. These segmentation differences account for the large gap between CER (5.9\%) and WER (85.7\%).

This pattern where models capture phonetic sequences more accurately than word boundaries is consistent with the challenges posed by morphologically rich languages, where agglutination and compounding are frequent.

\subsection{Sensitivity to Metric Weighting}

To understand how metric choice affects ranking and interpretations, we analyzed performance under different weightings of WER \& CER  (Table \ref{tab:sensitivity}). While relative rankings shift modestly, the fundamental finding is consistent.

\begin{table}[!ht]
    \centering
    \small
    \begin{tabular}{cccc}
    \toprule
    \textbf{WER Weight} & \textbf{CER Weight} & \textbf{Best Model} & \textbf{Score} \\
    \midrule
    75\% & 25\% & djelia/asr-v2 & 39.02\% \\
    50\% & 50\% & djelia/asr-v2 & 29.73\% \\
    25\% & 75\% & djelia/asr-v1 & 21.89\% \\
    100\% & 0\% & soloba-0.6b-v3 & 46.76\% \\
    0\% & 100\% & djelia/asr-v1 & 13.00\% \\
    \bottomrule
    \end{tabular}
    \caption{Best model performance under different WER/CER weightings.}
    \label{tab:sensitivity}
\end{table}

\section{Discussion}
\label{sec:disc}

\subsection{The Gap to Production Readiness}

Our results indicate that Bambara ASR is not yet ready for production deployment. To contextualize this gap, consider typical production requirements across different applications:

\textbf{Transcription services} (podcasts, meetings, legal proceedings) typically require WER below 10\%. Current Bambara systems are 35--40 percentage points above this threshold.

\textbf{Voice assistants and interactive systems} typically require a WER below 15\% to achieve an acceptable user experience. Current systems would result in nearly one in two words being misrecognized.

\textbf{Accessibility applications} (captioning, hearing assistance) have stringent accuracy requirements that current systems cannot meet.

\textbf{Voice-to-text input} requires near-perfect transcription for practical utility. At 47\% WER, correction effort may exceed that of manual typing.

These comparisons suggest that Bambara ASR currently requires significant further development before deployment in user-facing applications.

\subsection{Contributing Factors}

Several factors contribute to current performance levels, consistent with the challenges identified in recent systematic reviews of ASR in African language \cite{imam2025asr_slr}:

\paragraph{Limited training data.} Bambara remains a low-resource language in terms of labeled speech corpora. Although recent data collection initiatives have expanded data availability \cite{diarra2025dealinghardfactslowresource}, the total amount of Bambara speech data remains much lower the scale typically needed for high-performance ASR systems. By contrast, large multilingual models such as Whisper, which benefit from extensive multilingual training data, show catastrophic performance with unseen, underrepresented languages as we demonstrate in \ref{subsec:model-specific}. Benchmarking studies indicate that competitive ASR performance generally requires substantial volumes of labeled data \cite{nahabwe2025benchmarking}.

\paragraph{Domain mismatch.} Most available Bambara speech datasets consist of over-simplified spontaneous speech with limited vocabulary, recorded under controlled conditions (\citealp{diarra2025dealinghardfactslowresource}; \citealp{Diarra2022Griots}). This creates distribution mismatch when models encounter highly formal or inversely very informal registers, specialized vocabulary, or challenging acoustic conditions \cite{tall_2025_17672774}. Our benchmark also exposes this gap through its legal/constitutional domain.

\paragraph{Orthographic and dialectal variation.} Standardizing written Bambara is a recent research (\citealp{konta2014}; \citealp{vydrin:halshs-03909864}), despite the creation of a dedicated institution ---the Académie Malienne des Langues (AMALAN)--- the most recent orthography is not universally adopted, and dialectal variation across regions introduces additional complexity \cite{imam2025asr_slr}. Additionally, Bambara text available on the internet often features inconsistencies, old and mixed standards, models trained on one variant may struggle with others, fragmenting an already limited data pool.

\paragraph{Morphological complexity.} Bambara's agglutinative morphology makes word boundary detection inherently challenging. The gap between CER and WER across models reflects this difficulty phonetic patterns are captured more successfully than word structure.




\subsection{Implications for Research and Development}

Our findings have several implications:

\paragraph{Standardized benchmarking supports progress.} The field benefits from rigorous evaluation against common benchmarks. We encourage researchers to report results on standardized test sets in addition to internal evaluations.

\paragraph{Data collection should prioritize diversity.} Current data collection efforts, while valuable, may not adequately prepare models for real-world deployment. Future efforts should consider naturalistic speech, code-switching, dialectal variation, and varied acoustic conditions.

\paragraph{Architecture research may be needed.} The consistent underperformance of scaled multilingual models suggests that existing architectures may not be optimally suited to low-resource scenarios. Research into architectures designed for data-scarce settings may prove valuable.

\paragraph{Multilingual transfer has limits.} The poor performance of Whisper and similar systems demonstrates that multilingual pre-training does not automatically transfer to underrepresented languages. The dominance of RobotsMali's monolingual models suggests that, for Bambara and similar languages, targeted development appears more effective than relying on transfer from massive multilingual training.

\subsection{Directions for Progress}

Despite current limitations, our results suggest promising directions:

The success of smaller, Bambara-specific models (114M--600M parameters) over massive multilingual systems indicates that focused development yields better results than scale alone. The narrowing gap between proprietary and open-source solutions suggests that community-driven development can produce competitive systems. The reasonable CER performance (13--15\% for top models) indicates that phonetic modeling is more tractable than word-level transcription, suggesting that improvements in language modeling and vocabulary handling through post-processing could yield significant gains.

Closing the gap to production readiness will require sustained investment in data collection, architecture research, and evaluation infrastructure at scales that do not currently exist for Bambara and similar languages.

\section{Conclusion}
\label{sec:conclusion}

We present the first standardized benchmark for evaluating Bambara Automatic Speech Recognition systems and provide an empirical answer to the question posed in our title: \textbf{current Bambara ASR systems are not yet ready for production deployment.}

Our evaluation of 37 ASR models on a one-hour, studio-quality benchmark reveals that:

\begin{itemize}
\item The best-performing model on our benchmark is \textbf{djelia/asr-v2}, achieving a Combined Score of 29.73 (WER 47.50\%, CER 13.56\%) under ideal conditions.
\item No evaluated system reaches the 5--15\% WER range typical of production-ready ASR systems.
\item All OpenAI Whisper variants and commercial multilingual systems (not trained on Bambara) exhibit catastrophic failure, with WER exceeding 100\%, worse than how a randomly initialized model would perform. Suggesting that transfer learning fails where similarity between the target language and training languages stops.
\end{itemize}

These results should inform expectations for Bambara ASR deployment. Current systems may be suitable for research and development purposes, but deployment in production applications where users depend on accurate transcription should be approached with caution.

The benchmark and leaderboard are publicly available to support continued development and enable rigorous comparison of future systems. We hope this resource contributes to honest assessment of progress and motivates the sustained investment necessary to achieve production-ready Bambara ASR.

\section{Limitations}
\label{sec:limitations}

This benchmark has several limitations:

\paragraph{Simplified evaluation conditions.} Our benchmark represents near-ideal acoustic conditions: studio recording, professional speaker, high SNR, standardized orthography. Although we do speculate that the metrics reported here likely represent upper bounds on real-world performance, this assertion may not hold if some of the models that we evaluate have been trained on more naturalistic data. In other terms, the inverse assertion that models trained on natural data may experience more struggle on this benchmark may also be a valid interpretation. 

\paragraph{Single speaker and domain.} The current version features recordings from a single adult male speaker reading constitutional text. This limits assessment of speaker and domain variability, though it also provides a consistent and controlled evaluation environment.

\paragraph{Limited size.} One hour of audio is a minimal benchmark. However, consistent patterns across 37 models suggest findings would generalize to larger evaluations.

\paragraph{Metric limitations.} WER and CER may not optimally capture transcription quality for morphologically rich languages. Future work could explore morpheme-level metrics or semantic similarity measures.

\paragraph{Normalization sensitivity.} Our evaluation applied minimal text normalization (lowercase, punctuation removal, whitespace normalization) to ensure fair comparison. However, Bambara orthography permits substantial valid variation that our normalization does not fully address. Contractions such as \textit{b'a} versus \textit{b\textipa{E} a}, or the ambiguous \textit{k'a} which can legitimately expand to \textit{ka a}, \textit{k\textipa{E} a}, or \textit{ko a} depending on grammatical context, represent equivalent transcriptions that would be penalized as errors under standard WER computation. Similarly, compound word segmentation (\textit{y\textipa{E}r\textipa{E}mah\textipa{O}r\textipa{O}nya} versus \textit{y\textipa{E}r\textipa{E}ma h\textipa{O}r\textipa{O}nya}) and legacy orthographic variants (\textit{è}/\textit{\textipa{E}}, \textit{ny}/\textit{\textipa{\textltailn}}) introduce scoring artifacts unrelated to recognition accuracy. A more sophisticated normalization framework that accounts for these linguistic equivalences could yield different and potentially more meaningful error rates. Future work should investigate normalization strategies that distinguish genuine recognition errors from valid or outdated orthographic variation.

\paragraph{Code-switching.} Real Bambara speech frequently incorporates French, particularly in urban context but also formal settings, quite frequently. However, this first benchmark does not inform on a model ability to handle code-switching as this feature is deliberately absent from the data.

We view this benchmark as a foundation for continued development, with future versions incorporating speaker diversity, domain variation, naturalistic speech, and code-switching.

\section*{Data and Code Availability}

The benchmark dataset, evaluation code, and public leaderboard are available to support reproducibility and future research:

Benchmark Dataset : 
\begin{itemize}
    \item \url{https://huggingface.co/datasets/MALIBA-AI/bambara-asr-benchmark}
\end{itemize}

Public Leaderboard : 

\begin{itemize}
    \item \url{https://huggingface.co/spaces/MALIBA-AI/bambara-asr-leaderboard}
    \item \url{https://github.com/MALIBA-AI/bambara-asr-leaderboard}
\end{itemize}

\noindent We encourage researchers to submit their model results to the leaderboard and to report performance on this benchmark in future publications.

\bibliography{custom}

\end{document}